# Two-Stage Violence Detection Using ViTPose and Classification Models at Smart Airports

İrem Üstek, Jay Desai, Iván López Torrecillas, Sofiane Abadou, Jinjie Wang, Quentin Fever, Sandhya Rani Kasthuri, Yang Xing, Weisi Guo, Antonios Tsourdos

*Abstract*—This study introduces an innovative violence detection framework tailored to the unique requirements of smart airports, where prompt responses to violent situations are crucial. The proposed framework harnesses the power of ViTPose for human pose estimation and employs a CNN-BiLSTM network to analyse spatial and temporal information within keypoints sequences, enabling the accurate classification of violent behaviour in real-time. Seamlessly integrated within the SAAB's SAFE (Situational Awareness for Enhanced Security) framework, the solution underwent integrated testing to ensure robust performance in real-world scenarios. The AIRTLab dataset, characterized by its high video quality and relevance to surveillance scenarios, is utilized in this study to enhance the model's accuracy and mitigate false positives. As airports face increased foot traffic in the post-pandemic era, the implementation of AI-driven violence detection systems, such as the one proposed, is paramount for improving security, expediting response times, and promoting data-informed decision-making. The implementation of this framework not only diminishes the probability of violent events but also assists surveillance teams in effectively addressing potential threats, ultimately fostering a more secure and protected aviation sector. Codes are available at: https://github.com/Asami-1/GDP.

## I. INTRODUCTION

The aviation industry is highly sensitive to safety and timeliness, and airports face unique challenges in the violent events due to security and customs barriers that can complicate evacuations [1]. While surveillance can assist in investigating violent events, the low proportion of abnormal frames in all video frames makes it particularly challenging to detect violent events manually. Smart airports are implementing advanced technologies to enhance operational efficiency and intelligence [2]. AI-based violence detection systems can identify threats in real-time, accelerate response times to emergencies, and facilitate data-driven decision-making regarding resource allocation, which contribute to reducing labour costs and preventing violent incidents in smart airport [3, 4].

The goal of violence detection is to automatically and accurately identify violent events [3]. However, the subjectivity of violence's definition poses a challenge in translating the concept into mathematical expressions or computer-understandable language. Past research has used feature descriptors based on low-level features such as gradients, optical flows, and intensities [4]. The accuracy of these methods depends on the selected feature descriptor and researchers' understanding of violence. In contrast, image-based deep learning methods can automatically learn the rules of violent behaviour to adapt to different application scenarios and identify more complex violent behaviour. Additionally, deep learning-based techniques exhibit good accuracy, real-time performance, scalability, and transferability, which are advantageous for the smooth deployment and end-to-end optimization of violence detection systems in intelligent airports [5, 6].

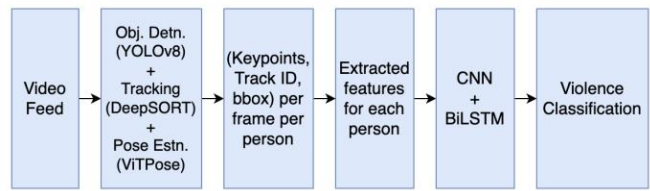

Fig. 1: Architecture of the proposed violence detection framework

Therefore, to enhance airport security, this study aims to develop a two-stage violence detection framework to estimate human posture and detect violent behaviour in real-time surveillance videos. This framework uses the pre-trained ViTPose model to detect human posture in each frame, preprocesses and extracts features from keypoints information, and then inputs these features into the CNN-BiLSTM network. This network further extracts features from keypoints information and learns the temporal information in keypoints sequences for violence detection. By identifying violence in real time, this framework can reduce the likelihood of airport violence incidents and assist airport surveillance teams in making effective responses when such events occur.

The main contribution of this study is the use of a keypoint-based approach for violence classification, utilizing the state-of-the-art ViTPose model for pose estimation instead of directly extracting features from images (which may be high-dimensional and abstract). This approach simplifies feature extraction, reduces the sample size and computational resources required for training and deployment, and accelerates model training, inference, and the entire pipeline's execution. Additionally, the use of key points reduces data size, decreases storage burden, improves operation speed, and facilitates implementation and deployment optimization. By focusing solely on human posture and movement information, this method minimizes interference caused by changes in image backgrounds and human body shape differences, while also reducing the influence of differences between datasets [7].

The authors are with the School of Aerospace, Transport, and Manufacturing, Cranfield University, UK, MK43 0AL. email {Irem.Ustek.197, Jay.Desai.336, I.LopezTorrecillas.352, Sofiane.Abadou.305, Jinjie.Wang.647, Quentin.Fever.387, Sandhyarani.Kasthuri.167, Yang.X, Weisi.Guo, A.Tsourdo}@cranfield.ac.uk

## II. RELATED WORKS

### A. Pose Estimation

Human pose estimation involves detecting the joints or keypoints of an individual from 2D or 3D images and videos [8]. There are classic methods for human pose estimation like HRNet, DCPose, and AlphaPose. HRNet can estimate high-resolution 2D or 3D poses but requires a lot of resources [9]. DCPose can detect and track human poses in real time with high efficiency and robustness [10]. AlphaPose combines CNN and a bottom-up method to estimate the key points of multiple people in real-time [11]. Transformers have recently been applied to pose estimation in computer vision, with ViTPose+ being the current state-of-the-art method. It uses a simple visual transformer to extract features from input images and introduces expert blending in the backbone network to improve performance. ViTPose+ has achieved new records on four challenging benchmarks while maintaining the same inference speed as ViTPose, which uses single-instance detection and has lower computational costs [12].

### B. Violence Detection

Violence detection is an application of Human Activity Recognition (HAR) in computer vision, which aims to accurately identify human actions from sensor data by analysing their spatiotemporal features. This study focuses on using human body keypoints to train a classifier, which acts as an attention mechanism [6]. Deep learning based models offer promising solutions because they can automatically learn and extract spatiotemporal features from input. Two main approaches are commonly used for HAR: end-to-end 3D CNN and the two-stream architecture using RNNs [13].

The 3D CNN technique involves using 3D convolutional kernels to analyse a video's spatiotemporal information, enabling the model to learn features from it. Previous research has successfully used this method as an end-to-end architecture [14, 15]. However, 3D CNN requires a significant amount of computing resources and time to train and infer, making it unsuitable for some real-time applications. The CNN+RNN approach involves using separate neural networks for spatial and temporal information processing. RNNs can model temporal dynamics and dependencies between frames but may suffer from the vanishing gradient problem, which can be addressed using LSTMs. Previous models achieved high accuracy using CNNs as spatial feature extractors and LSTMs for temporal information processing [16, 17].

## III. METHODOLOGY

### A. Datasets

Various datasets exist for pose estimation and violence detection individually. However, no common dataset is available, including ground truth for pose estimation and violence detection. Hence, we considered various separate datasets for each task and then combined them in a sequence to achieve the ultimate objective of violence detection using pose estimation.

#### 1) Dataset Selection

There are various datasets available for training pose estimation models. We used a pre-trained model (ViTPose), demonstrating state-of-the-art results on datasets such as COCO, AIC (AI Challenger), MPII, CrowdPose, and WholeBody for human pose estimation.

We analysed widely used RWF-2000, UCF-Crime, and also AIRTLab dataset. The RWF-2000 dataset comprises 2,000 real-world video clips, evenly split between violent and non-violent content, sourced from movies, surveillance cameras, and social media [18]. The UCF-Crime dataset is a large-scale collection of real-world surveillance videos featuring 13 types of crime and regular activities, such as fighting, burglary, and vandalism [19]. Lastly, the AIRTLab dataset, created by the University of Rome III's Artificial Intelligence and Robotics Laboratory, consists of 350 MP4 video clips at 1920 x 1080 resolution and 30 fps, varying in duration from 2 to 14 seconds, designed specifically for automatic violence detection [20]. In AIRTLab Dataset, the clips are annotated as either "violent" or "non-violent" based on the presence or absence of physical aggression. The AIRTLab dataset is characterized by its high video quality, an essential feature for accurate and reliable human pose estimation. This dataset has been selected to enhance the accuracy of our model by providing more precise keypoints data. By utilizing the AIRTLab dataset, we aimed to improve the performance of our model and make it more effective in real-world scenarios. Due to these reasons, this dataset is selected.

#### 2) Dataset Preprocessing

##### a) Manual Annotation

To achieve accurate violence recognition, a frame-level manual annotation was conducted on the dataset, categorizing individuals as neutral, aggressor, or victim. Noise-contributing boundary boxes, resulting from false detections by YOLO or discontinuous human behaviour, were eliminated to improve training data quality. Misidentified boundary boxes by DeepSort were merged, and individuals' behavioural categories were labelled by evaluating their behaviour in continuous frames. Fig. 1 shows the output of our AI pipeline. Fig. 3, Fig. 4 and Fig. 5 show examples of manual labelling scenarios.

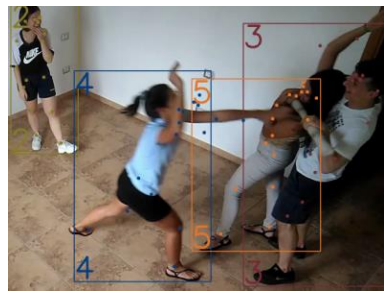

Fig. 2: An example frame after the three (YOLO, DeepSort and ViTPose) models have been applied.

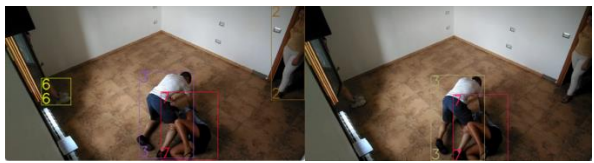

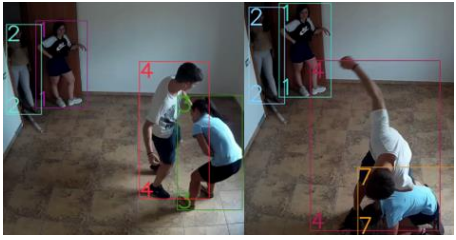

Fig. 3: An example of removing unnecessary track IDs with Python script (IDs 2 and 6 removed).

Fig. 4: An example frame having different IDs for the same person across frames.

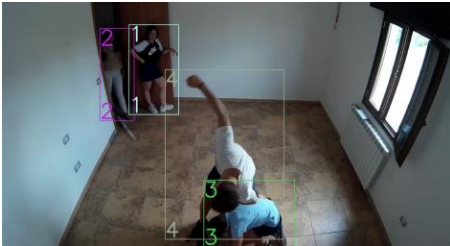

Fig. 5: An example of merging the IDs for the same person (IDs 3 and 7 are merged).

*b) Feature Engineering*

In this study, we used two feature engineering strategies from pose estimation for the classifier: distance calculation and angle calculation. Distance-based features detect violent behaviour by assessing deviations from the body's normal state, such as leaning, tilting, or limb movements. Angle-based features utilize angles between body keypoints to provide information about body orientation and movement, aiding in violent and non-violent behaviour classification. After comparing the two strategies, we found that distance-based approaches had better accuracy and selected models accordingly.

*c) Extracting Each Individual*

This study trains AI models using individual keypoints data to enhance robustness and generalization by capturing unique behavioural patterns while avoiding cross-interference. Despite requiring higher quality datasets and potentially affecting real-time operation speed, this approach is preferable given the study's focus on identifying aggressors and victims in violent events. Training with multiple individuals' keypoints data introduces additional challenges, such as increased model complexity, mutual interference, and potential information loss due to dimension reduction. Therefore, using individual body keypoints data for model training is adopted to meet study requirements and ensure higher recognition accuracy while maintaining a coherent methodology.

*d) Data Augmentation*

We compared AI model performance using two-time intervals: 1-second and 2-second intervals. The 2-second interval provided more information per data point but generated fewer data points, while the 1-second interval created more data points with less information. Using the Windows approach to analyse time series data, we broke the data into smaller windows to identify violent behaviour patterns. Generating more data points with smaller window sizes enabled more frequent data sampling, providing a clearer view of the data and helping identify subtle behavioural changes. This approach improves the classifier's ability to accurately identify violent and non-violent behaviours, aiding in the development of effective interventions.

### B. Tracking and Object Detection

To effectively carry out the pose estimation and classification, it is necessary to previously detect the bounding boxes of each person and track them through the video. While considering different methods and pre-trained models, it is important to obtain good accuracy while keeping inference time low, since the main objective is to deploy in a real-time application. In our work, two subsystems have been used separately: for the detection of people, it has been decided to use the smallest pretrained model of YOLOv8 (nano) that offers the fastest inference time with enough accuracy for this study; on the other hand, DeepSORT is used for tracking people across multiple frames. This method combines the classic Simple Online and Realtime Tracking (SORT) algorithm with a deep learning model which helps to reduce tracking errors due to missed detections or bounding box occlusion.

### C. Pose Estimation by ViTPose

As in this study people body keypoints are used to identify and classify possible violent behaviours, one of the main systems in the AI engine pipeline is the pose estimator itself. ViTPose is an algorithm that combines the strengths of Vision Transformers (ViT) and pose estimation techniques to achieve high-precision human pose estimation in images or videos. Vision Transformers divide an input image into smaller patches and process them using a trans-former architecture, while pose estimation focuses on identifying the spatial locations of key body parts in the images or videos. Once again, being inference time the deciding factor to choose a pretrained model, ViTPose+-S was implemented in the study.

### D. Classification

As mentioned earlier, the output of the pose estimation along with the person bounding box track ID generated from previous subsystems, are preprocessed to extract the desired features (either distance or angle based) from each individual in a set duration sequence (one or two seconds long). These sequences are the input of the classification model.

In this study, four different model architectures were considered: on the one hand, Long Short-Term Memory (LSTM) networks which are Recurrent Neural Networks (RNN) specifically designed to address the vanishing gradient problem and effectively capture and store long-range dependencies within a data sequence; on the other hand, Bidirectional Long Short-Term Memory (BiLSTM) networks which are an extension of LSTMs that operate in two directions, allowing the model to learn from past and future con-texts within a sequence; finally, a variation of the two previous networks mentioned having Convolutional Neural Network (CNN) layers as the first layer of the architecture acting as feature extractors, capturing spatial information and local patterns within the input data. During training, CNN,

LSTM and BiLSTM layers were followed by Dropout layers. Additionally, a custom callback early stopped the training after 50 epochs with no improvement to avoid overfitting.

All four model architectures were trained in both distance-based and angle-based features and in 1-second and 2-second-long sequences. Furthermore, the models were hyperparameter tuned for different fixed values of dropout rates, Batch size, number of LSTM Layers in the architecture and Learning rate using SGD and Adam Optimizers.

### E. Integration to SAFE

The work was conducted in collaboration with SAAB and DARTeC (Digital Aviation Research and Technology Centre) in order to integrate our AI model into one of their applications. SAFE is an open-integration software platform for mission-critical operations. It is utilized in command-and-control rooms in various high-security areas such as airports, law enforcement operations centres, and maximum security correctional facilities. The goal was integrating the developed AI model into the SAFE environment, allowing monitoring of its predictions in one place. We integrated our AI model into an AI Engine. The engine was built using Flask, a python-based lightweight and modular framework that allows creating of applications interacting with AI technologies easily. We also built a docker container for the application to make it deployable regardless of the environment. We then integrated this engine into the SAFE environment. To communicate with SAFE's application server, we used Kafka, a distributed streaming platform that offers high scalability, durability, and flexibility to build real-time streaming applications. The workflow is shown in Fig. 6.

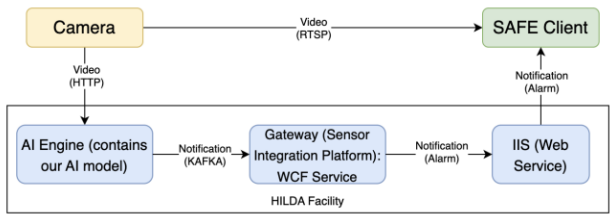

Fig. 6: System Architecture

*1) Sending video feed to engine and client:* To perform Violence Detection, video feeds are captured from these cameras and given to the AI engine as well as the SAFE client terminal directly.

*2) AI engine inference:* AI engine receives the video feed from the camera system and performs Violence Detection on it. A block diagram for the same is given in Fig. 1.

*3) Notifying Gateway and SAFE client:* Gateway is used to consume the KAFKA message given by the AI engine and forward it to SAFE's client terminal for further display and alerts. All alerts for violence can be shown in real-time on this layout, and the operator can then take appropriate action.

## IV. MODEL EVALUATION

### A. Classification

The study employed both angle-based and distance-based features derived from raw keypoint datasets with sequence lengths of 1-second and 2-seconds to train 2592 models (162 hyperparameter combinations x 4 different dataset features x 4 different model types). Results were obtained in terms of average and maximum accuracy metrics and training time. It is observed that the distance-based features dataset performed better compared to the angle-based features dataset for both 1-sec and 2-secs of sequences, therefore, further analysis is only done through the distance-based feature dataset.

### B. Comparison of the Best Models

After conducting a comprehensive analysis of various models using multiple datasets on 2592 models, several promising models were identified that meet specific evaluation criteria. The assessment included analysing the models' train accuracy, test and validation accuracy, and examining the loss and accuracy plots to ensure that they are not overfitted. In addition, the real-time performance of each model was evaluated using the AI Engine application to detect violent behaviour in camera footage. It can be concluded that valuable information about violent behaviour can be extracted using distance features of body keypoints.

TABLE I. COMPARISON OF THE TWO CANDIDATE BEST MODELS FOR MULTICLASS CLASSIFICATION

| Model | Dataset | Train Dataset Size | Train Accuracy | Test Accuracy |
|---|---|---|---|---|
| CNN-BiLSTM Model (A) | Distance-based & 1 second sequence | (8048, 10, 24) | 0.813 | 0.798 |
| CNN-BiLSTM Model (B) | Distance-based & 2 second sequence | (6558, 20, 24) | 0.853 | 0.806 |

Model (B), trained on a distance-based dataset with 2-second sequences, achieved the highest train accuracy (0.853) and test accuracy (0.806) among several models. Model (A), trained on a standard 1-second dataset, performed closely to Model (B) in terms of test accuracy (TABLE I). However, in real-time classification, Model (A) was observed to outperform Model (B), which is important for the study's goal of real-time violence detection in live camera feeds (Fig. 10).

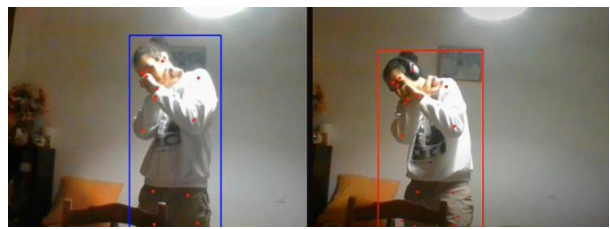

Fig. 7: Real-time performance comparison of Model (A) (on the left) and Model (B) (on the right) from the AI Engine (Blue bounding box: Victim, Red bounding box: Aggressor, Green bounding box: Neutral).

Model (B) struggles to consistently detect the person in Fig. 7 as a Victim, unlike Model (A). This performance difference can be explained by dataset size and sequence length. Model (A) was trained on a larger dataset (8048 samples) and a shorter sequence length of 1 second, which may have provided more diverse and representative examples, leading to better generalization and sensitivity in capturing real-time violence cues. Therefore, Model (A) was better suited for real-time violence classification from live camera feed and was selected as the preferred choice for the final model.

### C. Final Model Results

Based on the analysis of model results, the CNN-BiLSTM model utilizing a distance-based approach and a dataset with 1-second sequences, Model (A) (1 CNN layer, 5 BiLSTM layers with dropouts and a final densely connected output layer with softmax activation), has been identified as the optimal choice for multiclass classification. The model was trained using a training dataset size of (8048, 10, 24), a learning rate of 0.1, batch size of 64, dropout rate of 0.4, and the stochastic gradient descent (SGD) optimizer. The model achieved a train accuracy of 0.813 and a test accuracy of 0.798 at the best epoch (114).

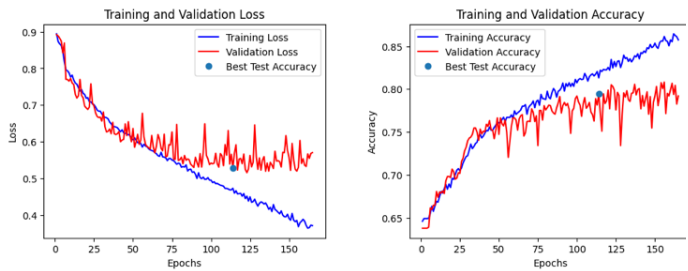

Fig. 8: Loss and accuracy plots of final model in train and validation datasets

TABLE II. CLASSIFICATION REPORT OF THE FINAL MODEL (MODEL (A))

|  | Precision | Recall | F1-score | Support |
|---|---|---|---|---|
| **Neutral** | 0.86 | 0.91 | 0.89 | 1465 |
| **Aggressive** | 0.59 | 0.61 | 0.60 | 376 |
| **Victim** | 0.52 | 0.28 | 0.36 | 205 |

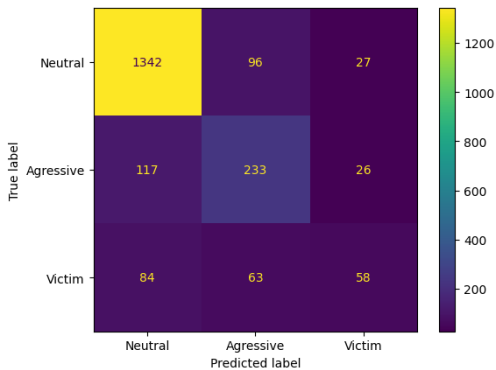

Fig. 9: Confusion matrix for multi-class violent behaviour classification.

The model's loss and accuracy plots indicate that it achieved its highest test accuracy around the 115th epoch while preventing overfitting using a custom callback (Fig. 8). The F1-score, which provides a balanced measure of both precision and recall, was used to evaluate the model's performance, and it showed that the 'Neutral' class had the highest F1-score of 0.89, while the 'Victim' class had the lowest F1-score of 0.36 (TABLE II). However, this could be attributed to the smaller number of instances in the 'Victim' class compared to other classes. Overall, this final model is utilized in the AI engine to detect violent behaviour in real-time by classifying individuals as Aggressor, Victim, and Neutral.

### D. SAFE Interface Development and Demonstration

We deployed our development model in HILDA, received frames from DARTeC's camera to AI engine for inference, and classified violent behaviour in the video. Fig. 10 shows the output of our model's classification on DARTeC's camera. Our model managed to classify aggressors and victims adequately. Our model achieves a total real-time classification speed of 0.1 seconds, allowing for accurate processing of video streams at 10 frames per second, which has proven sufficient for achieving accurate classification.

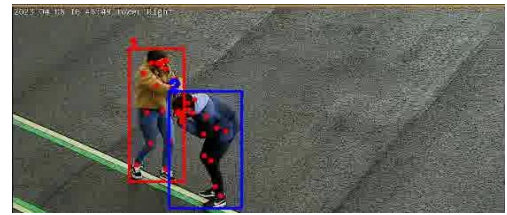

Fig. 10: Output of the model on DARTeC's camera feed.

Fig. 11 shows SAFE's layout, we can see that our KAFKA messages were successfully received, alarms have been activated on the client's side.

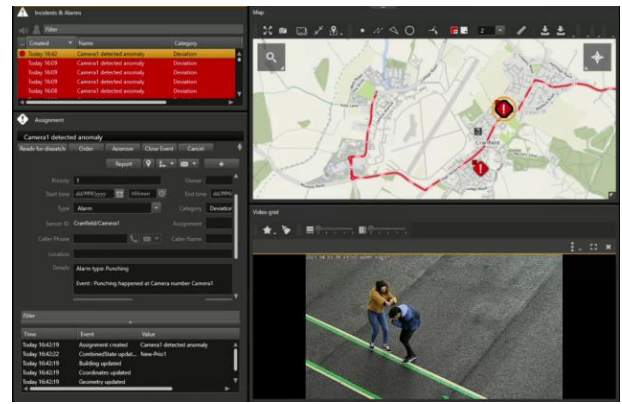

Fig. 11: SAFE user-interface during inference of the model

## V. FUTURE WORKS

The proposed activity recognition framework shows promise, but there are potential avenues for future work to enhance its performance and adaptability. Incorporating attention mechanisms into the classifier can improve the model's ability to capture temporal relationships between keypoints, potentially boosting performance. Additionally, replacing traditional AI models (CNN, LSTM, BiLSTM) with transformer models can handle long-range dependencies and parallel computation, making them suitable for tasks with large input data. Transformer models' parallelization capabilities may also speed up training and inference, making the framework more scalable for real-time applications. Increasing data diversity through multiple datasets and applying data augmentation techniques can improve performance, especially in cases where the sample size is small, or behavior type distribution is imbalanced. Moreover, considering camera calibration and normalization can ensure effective generalization across different angles and scenes, enhancing the framework's robustness and adaptability.

VI. CONCLUSION

In conclusion, the implementation of a violence detection model employing a two-stage approach pose estimation and violence classification provides an effective solution for detecting violent behaviour in real time. This system is particularly beneficial in high-security environments such as smart airports, where ensuring public safety is of importance. The success of this approach is primarily attributed to the use of ViTPose for pose estimation, the CNN-BiLSTM model for violence classification, and the carefully curated AIRTLab dataset, which minimizes false positives. To ensure the practicality and effectiveness of the system, end-to-end testing was conducted at the DARTeC building at Cranfield University, incorporating live feed from the camera and real-time alert notifications on the SAFE client. This comprehensive testing demonstrated the model's ability to detect violence accurately and efficiently in real-world environment.

ACKNOWLEDGEMENT

The authors appreciate the support of the SAAB SAFE team, including Wikander Andreas and Henningsson Andreas, for integrating the SAFE environment. They also acknowledge George Yazigi for managing computing resources at Cranfield DARTeC.